\documentclass[conference]{IEEEtran}
\IEEEoverridecommandlockouts
\usepackage{cite}
\usepackage{amsmath,amssymb,amsfonts}
\usepackage{algorithmic}
\usepackage{textcomp}
\usepackage{xcolor}
\usepackage{enumerate}
\usepackage{array}
\usepackage{stfloats}
\usepackage{url}
\usepackage{verbatim}
\usepackage{graphicx}
\usepackage{subcaption}
\usepackage{diagbox}
\usepackage{booktabs}
\usepackage{multirow}
\usepackage{float}
\usepackage{bm}
\usepackage{hyperref}
\graphicspath{{./figures/}}

\def\BibTeX{{\rm B\kern-.05em{\sc i\kern-.025em b}\kern-.08em
T\kern-.1667em\lower.7ex\hbox{E}\kern-.125emX}}
\begin{document}


    \title{Deep Learning for Biopsy-Free Subtyping of Basal Cell Carcinoma from Dermatoscopic Images
    \thanks{This work was funded by the Special Account for Research
    Funds of the Aristotle University of Thessaloniki (90169/2024)
        }
    }
    \author{
        \IEEEauthorblockN{
            Alexandros Papadopoulos,
            Chrysa Episkopou, \\
            Ioannis Sarafis
            and Anastasios Delopoulos
        }
        \IEEEauthorblockA{
            \textit{School of Electrical and Computer Engineering} \\
            \textit{Aristotle University of Thessaloniki, Greece}\\
            \{palexang, cepiskop, iosarafis, antelopo\}@ece.auth.gr
        }
        \and
        \IEEEauthorblockN{
            Aimilios Lallas
        }
        \IEEEauthorblockA{
            \textit{First Department of Dermatology} \\
            \textit{Aristotle University of Thessaloniki, Greece}\\
            alallas@auth.gr
        }
    }

    \maketitle

    \begin{abstract}
      Basal Cell Carcinoma (BCC) is the most common type of skin cancer, accounting
      for nearly 80\% of skin cancer diagnoses. Its optimal clinical management
      is guided by the distinct histopathologic subtype, with
      aggressive variants requiring more drastic measures. In current clinical
      practice, subtyping relies on skin biopsies, a procedure
      both costly and invasive. In this paper, we conduct a preliminary
      investigation into using deep learning for BCC subtyping,
      solely from a single dermatoscopic image of the
      lesion. Given the limited data at our disposal, we employ pre-trained vision transformers (ViTs),
      a state-of-the-art family of models highly effective for challenging downstream tasks with limited labeled data. Through repeated stratified k-fold
      cross-validation, we demonstrate that ViTs can achieve superior
      performance (AUC 0.784 on a dataset of 1271 dermatoscopic images of various BCC subtypes) over standard CNN-based baselines as well as previously-reported
      human reader performance, on the task of differentiating aggressive BCCs from other subtype families. These initial findings highlight the
      potential of combining deep learning and dermatoscopy to
      provide a biopsy-free alternative for BCC subtyping, thus aiding in
      improving treatment planning and patient outcomes.

    \end{abstract}

    \IEEEpeerreviewmaketitle


    \section{Introduction}
    Basal Cell Carcinoma (BCC) is a type of skin cancer caused primarily by
    ultraviolet radiation (UV). As such, it most often develops on body
    parts exposed to the sun\cite{crowson2006basal}.
    Treatment options for BCC depend on its histopathologic subtype, with
    micrographic surgery suggested as the optimal treatment for
    aggressive subtypes, unlike non-aggressive and superficial tumors that may be
    treated more conservatively\cite{peris2023european}.
    Prior to any treatment however, the subtype must first be established by a diagnostic biopsy,
    a procedure both costly and invasive that may also cause significant patient discomfort.
    Some imaging techniques, like optical coherence tomography
    or reflectance confocal microscopy, coupled with expert human evaluators or
    deep learning models, have been proposed as a viable alternative
    to diagnostic biopsies\cite{chen2022deep, mandache2018basal}. However, only
    a few studies\cite{lallas2014accuracy, camela2024dermatoscopic, longo2014classifying}
    have examined whether dermatoscopy in particular, by far the most common and accessible imaging modality in the field, can aid in BCC subtyping.

    Dermatoscopy (also known as dermoscopy) is a non-invasive
    imaging technique widely used by dermatologists for the examination and diagnosis of skin
    lesions~\cite{kittler2011dermoscopy}. It involves a handheld device called a dermatoscope, which provides
    magnification (typically around $10\times$) and illumination through polarized or non-polarized
    light sources.
    Dermatoscopy enables clinicians to visualize subsurface structures and morphological features
    not visible to the naked eye, such as pigment networks, vascular patterns,
    globules, and crystalline structures\cite{argo2022dermoscopy}.
    It significantly improves diagnostic accuracy
    compared to visual examination alone, facilitating early detection and differentiation of benign and malignant
    skin conditions. Recent advancements\cite{magalhaes2024systematic} have combined dermatoscopy
    with deep learning-based AI systems, to further enhance diagnostic capabilities and support
    clinical decision-making.

    Deep learning has been a natural fit for automating medical image analysis,
    with many applications during the past decade, and especially so in dermatology, for classification of skin
    lesions, like melanoma\cite{9656127,lopez2017skin} and other skin cancers\cite{9121248, kassem2021machine}.
    Within an abundant literature, however, little attention has been given to BCC subtyping,
    an essential task for treatment planning as we have discussed.
    The existing works focus on using
    either \emph{whole-slide histopathology images} (WSI)\cite{yacob2023weakly, bunguardean2022deep},
    or other imaging techniques like \emph{reflectance confocal microscopy (RCM)}\cite{chen2022deep}
    and \emph{optical coherence tomography (OCT)} \cite{mandache2018basal}.
    To the best of our knowledge, however, no works have
    considered combining dermatoscopy with deep learning
    for subtyping basal cell carcinomas.

    Setting deep learning aside, few works have studied if dermatoscopy is informative enough
    to differentiate low-risk BCCs (superficial, nodular) from more aggressive subtypes (infiltrative, morpheaform, micronodular).
    The early work of\cite{lallas2014accuracy} identified and evaluated various dermoscopic criteria for differentiating
    superficial BCC (sBCC) from other subtypes, resulting in an algorithm with 81.9\% sensitivity
    and 81.8\% specificity in a dataset
    of 77 sBCCs and 258 non-sBCCs.
    A follow-up work\cite{camela2024dermatoscopic} investigated a more clinically relevant problem: whether human
    readers can accurately diagnose aggressive BCCs from a single dermoscopy image.
    To that end, 30 expert human evaluators with variable levels of
    experience were asked to perform an initial classification of 235 BCCs into superficial (sBCC), non-superficial (nsBCC) and aggressive
    (according to the same categorization described in Section\ref{subsec:bcc-subtype-dataset}).
    After a washout period of 2 months, they were explicitly instructed on the
    different dermatoscopic criteria for each type and repeated the classification
    of the 235 BCCs. The authors then compared the performance of the evaluators before
    and after the training and noticed substantial
    improvements in sensitivity (from $42.3\%$ to $56.4\%$) with
    a minor decrease in specificity (from $71.7\%$ to $70.7\%$).

    Here we make a first step towards filling the above gap in
    the literature. More specifically, we present
    an automated approach for differentiating aggressive from
    non-aggressive BCCs based on dermatoscopic images, a binary classification task identical
    to the one pursued in\cite{camela2024dermatoscopic}.
    To that end, we follow a data-driven approach based on deep learning and
    particularly vision transformers\cite{dosovitskiy2020image}, a choice
    dictated by the very small annotated BCC
    subtype dataset at our disposal (presented in Section \ref{subsec:bcc-subtype-dataset}), which
    makes it difficult to train a data-efficient deep learning model.
    In doing so, we take advantage of the unsupervised
    pre-training of vision transformers on vast amounts
    of data, which has been shown to be very effective in downstream tasks with limited labeled data\cite{bao2021beit}, like the learning problem we face here.

    This paper is structured as follows:
    In Section~\ref{sec:datasets}, we present the AUTh BCC subtype dataset which
    motivated and informed this work, along with any other datasets that were used for auxiliary purposes.
    In Section~\ref{sec:proposed-approach}, we outline the proposed deep learning approach
    for building an accurate aggressive BCC subtype predictor.
    In Section~\ref{sec:experimental-results}, we describe the experimental setup,
    outline the training details and present and discuss the experimental results.
    Finally, we offer concluding remarks in Section~\ref{sec:conclusions}.

    \section{Data}\label{sec:datasets}
    We begin with the presentation of the data at our disposal.
    First, we present the AUTh BCC subtype dataset which informed most of the design choices we made.
    We then briefly describe the ISIC archive, a public dermatoscopy dataset that was used for adapting our models to the dermatoscopy domain, prior to finetuning on our target task.

    \subsection{AUTh BCC subtype dataset}\label{subsec:bcc-subtype-dataset}
\begin{figure}
        \centering
        \includegraphics[width=.5\textwidth]{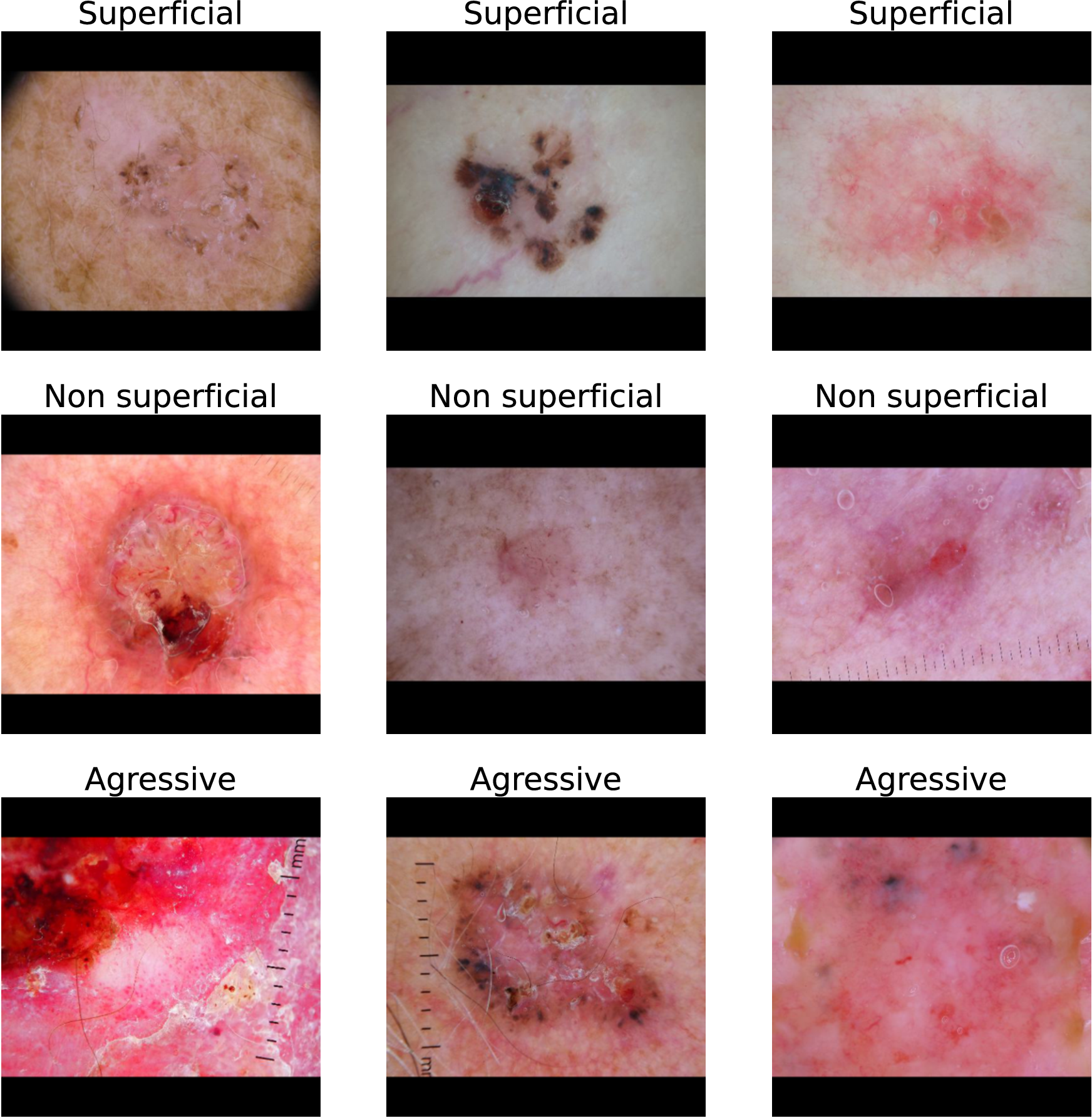}
        \caption{Sample images from the BCC subtype dataset used in this study.
        First row: Superficial BCC,
            Second row: Non-aggressive, non-superficial
            BCC, Third row: Aggressive BCC}
        \label{fig:data-sample}
    \end{figure}

    The AUTh BCC subtype dataset contains 1271 dermatoscopic images of
    Caucasian patients with BCC, diagnosed at the
    First Department of Dermatology, Aristotle University of Thessaloniki, Greece,
    between January 2018 and December 2022.
    The images were captured using Dermlite Foto
    equipment (3Gen, Dana Point, CA) at 10-fold magnification.
    Mixed histopathological subtypes and incompletely excised BCCs
    (shave or punch biopsy) were excluded. For each tumor, the
    biopsy-confirmed histopathological subtype was recorded and classified into three categories:
    \begin{enumerate}
        \item \emph{Superficial BCC (sBCC)}
        \item \emph{Non-aggressive, non-superficial BCC (nsBCC)}: nodular BCC, nodulocystic BCC, and Pinkus fibroepithelioma.
        \item \emph{Aggressive BCC}: infiltrative, morpheaform, micronodular, and metatypical BCC.
    \end{enumerate}
    Additional metadata like age and sex of the patient, as well as tumor location were also collected.
    The demographic characteristics of the dataset are given in Table~\ref{tab:bcc_demographics}, while
    a sample of the dataset can be seen in Figure~\ref{fig:data-sample}

    \begin{table}[h]
        \centering
        \caption{Demographic data and tumor characteristics of the AUTh BCC subtype dataset}
        \begin{tabular}{l|cc|c}
            \toprule
            & \multicolumn{2}{c|}{\underline{\textbf{Low Risk}}} & \underline{\textbf{High Risk}} \\
            & \textbf{Superficial} & \textbf{Non-superficial} & \textbf{Aggressive} \\
            & (n=275)              & (n=726)                  & (n=270)             \\
            \midrule
            \textbf{Age (Mean ± SD)}
            & 66.2 ± 14.3
            & 70.0 ± 13.1
            & 69.9 ± 12.8 \\
            \textbf{Sex (M / F / NA)}
            & 98 / 89 / 88
            & 327 / 212 / 187
            & 70 / 42 / 158 \\
            \midrule
            \textbf{Tumor location} &                      &                          &                     \\
            Head and neck
            & 69 (25.1\%)
            & 373 (51.4\%)
            & 83 (30.7\%) \\
            Trunk
            & 102 (37.1\%)
            & 136 (18.7\%)
            & 25 (9.3\%) \\
            Upper extremities
            & 11 (4.0\%)
            & 15 (2.1\%)
            & 2 (0.7\%) \\
            Lower extremities
            & 6 (2.2\%)
            & 12 (1.7\%)
            & 2 (0.7\%) \\
            NA
            & 87 (31.6\%)
            & 190 (26.2\%)
            & 158 (58.5\%) \\
            \midrule
            Total & \multicolumn{2}{c|}{1001} & 270 \\
            \bottomrule
        \end{tabular}
        \label{tab:bcc_demographics}
    \end{table}

    \subsection{ISIC archive}\label{subsec:isic-archive}
    The International Skin Imaging Collaboration (ISIC) Archive is
    an open-access repository of 503,955 publicly-available dermatology-related
    images, including dermatoscopic, RCM and whole-body photographs.
    For the purposes of this work, we used only the dermatoscopic images of the archive for
    which a lesion diagnosis was available.
    After discarding some severely under-represented classes,
    we ended up with the subset of Table~\ref{tab:isic_stats}.

    \begin{table}[!h]
        \centering

        \caption{Statistics of the subset of the publicly available ISIC dermatoscopy images that was used for adapting models pre-trained on natural images to the dermatoscopy domain.}
        \begin{tabular}{ll}
            \toprule
            Lesion                & Count  \\
            \midrule
            Actinic keratosis       & 1,367  \\
            Basal cell carcinoma    & 4,921  \\
            Benign keratosis        & 4,380  \\
            Dermatofibroma          & 420    \\
            Melanoma                & 7,349  \\
            Nevus                   & 32,697 \\
            Other                   & 972    \\
            Squamous cell carcinoma & 1,372  \\
            Vascular lesion         & 348    \\
            \midrule
            Total                   & 53,826 \\
            \bottomrule
        \end{tabular}
        \label{tab:isic_stats}
    \end{table}


    \section{Proposed approach}\label{sec:proposed-approach}

    \subsection{Problem statement}\label{subsec:problem-statement}
    Our goal is to build a deep learning model that will accurately
    differentiate the high-risk aggressive BCC subtype
    from other BCC subtypes considered to be low-risk,
    based on a single dermatoscopic image of the tumor.
    Given the complexity of the task (evidenced by the results of~\cite{longo2014classifying}) and the scarcity
    of labelled dermatoscopic images of BCC subtypes at our disposal, we
    opted for a vision transformer model, as the unsupervised pre-training of these models on vast amounts
    of data has rendered them very effective in downstream tasks with limited labeled data\cite{bao2021beit}.

    \subsection{Image pre-processing}\label{subsec:image-pre-processing}
    Since images in our dataset vary significantly in resolution due to the use of different
    dermatoscopic devices, we first apply a standardized preprocessing step to ensure consistency.
    Specifically, each image is padded to a square aspect ratio and subsequently resized to a fixed
    resolution of $512\times512$ pixels. While this resolution is higher
    than what is typically used in medical image analysis tasks, we deemed it to be a reasonable
    trade-off between preserving fine-grained image details that may be crucial for identifying subtle
    diagnostic features and maintaining computational feasibility.

    \subsection{Model architecture}
    For our vision transformer backbone, we employed the
    BEiT model\cite{bao2021beit}, a vision transformer
    architecture that achieved state-of-the-art performance in a number of benchmarks
    at the time of its release. BEiT is inspired by the
    success of BERT in natural language processing and adapts a similar masked
    image modeling strategy for vision tasks.
    During pre-training,
    BEiT learns to reconstruct discrete visual tokens (obtained from a
    pretrained tokenizer such as a VQ-VAE) from masked patches,
    enabling it to capture rich semantic representations of visual data in a highly efficient manner.

    Given the high resolution of our dermatoscopic images, we opted
    for the BEiT-Large variant, which operates on input images of
    size $512 \times 512$ with a patch size of $16 \times 16$.
    This configuration results in a grid of $32 \times 32$ input tokens and
    approximately 300 million parameters.
    The model consists of 24 transformer layers with 16 attention heads per layer and a
    hidden dimension of 1024.
    This large-scale architecture allows for substantial representational
    capacity, making it particularly suitable for modeling fine-grained visual distinctions such
    as those found between subtypes of Basal Cell Carcinoma.

    \subsection{Feature adaptation}\label{subsec:feature-adaptation}
    Since BEiT was originally pre-trained on natural image datasets,
    it is essential to first adapt it to the domain of medical imaging, specifically
    dermatoscopic images, before applying it to the downstream task of BCC
    subtyping. To facilitate the domain adaptation, we begin with the publicly available
    pre-trained weights of BEiT and conduct a fully supervised finetuning on the ISIC
    subset described in Table~\ref{tab:isic_stats}, which contains approximately 53,000 dermatoscopic
    images spanning 9 distinct classes of skin lesions.
    We finetune the model for 100 epochs using 80\% of the dataset for training, while reserving the
    remaining 20\% for monitoring the model’s generalization performance.
    After training, we select the checkpoint with the highest validation accuracy and use
    it as the initialization point for our subsequent BCC subtype classification model.
    This intermediate step
    ensures that the transformer learns domain-specific features relevant to dermatoscopic imagery, thereby
    improving its effectiveness on the target classification task.

    \subsection{Finetuning for BCC subtype classification}
    After feature adaptation, we proceed
    with finetuning the model to the downstream task of
    differentiating aggressive BCC from lower-risk subtypes.
    Following standard practice for BEiT, we employ a low initial learning
    rate of $10^{-5}$, combined with a weight decay factor of
    $10^{-4}$, and apply a cosine annealing schedule to
    gradually reduce the learning rate during training.
    The model is trained for a total of 10 epochs on a training subset of the AUTh BCC subtype dataset,
    using a batch size of
    16, while its performance is continuously monitored on an independent validation set to prevent overfitting.
    To account for the imbalance (see Table \ref{tab:bcc_demographics}) between the high- and low-risk classes, we perform oversampling of the minority class,
    so that both classes are represented equally in a training batch.
    We apply moderate data augmentation during
    training, including random vertical and horizontal flips, subtle brightness and contrast adjustments,
    as well as random shifts, scalings, and rotations. No augmentation is applied during the
    validation or testing phases. After training,
    the checkpoint with the highest validation performance is selected and subsequently evaluated on a completely
    independent test set to assess its ability to generalize to unseen data.
     More details
    on the training/validation/test splits are provided in Section \ref{sec:experimental-results}.

    \section{Experimental results}\label{sec:experimental-results}

    To allow for a reliable estimation of the model's generalization performance and owing
    to the small dataset
    size, we opted for a stratified 5-fold cross-validation scheme which we repeated three times.
    In each iteration of this repeated k-fold scheme, 20\% of the fold training set was
    set aside for validation. After training, the best performing model on the validation set
    (according to ROC-AUC) was evaluated on the fold test set.
    In addition to BEiT, we also experimented with two ResNet-based baseline models
    (trained with the same setup and computational budget) for comparison.
    We present the results
    of this experiment in Table~\ref{tab:results3}, alongside the previous human expert performance
    on the same task, as reported in~\cite{camela2024dermatoscopic}.
    In Table~\ref{tab:results3}, the reported metrics
    have been averaged across all training-test splits of the repeated k-fold scheme.
    For our models, sensitivity was calculated at the operating point yielding $0.70$
    specificity, to align with the reported specificity
    of human experts in \cite{camela2024dermatoscopic} and to facilitate comparisons with them.

    \begin{table}[!h]
        \centering
        \caption{Average performance metrics over 3 repetitions of a stratified 5-fold
        cross-validation experiment. For deep learning models,
            sensitivity was calculated at the operating point of 70\%
            specificity to allow comparisons with \cite{camela2024dermatoscopic}}
        \begin{tabular}{l|c|cc}
            \toprule
            \textbf{Method} & \textbf{AUC} &
            \textbf{Specificity} & \textbf{Sensitivity} \\ \midrule
            Human experts (pre)\cite{camela2024dermatoscopic}  & NA                       & 71.1$\pm$7.7  & 42.3$\pm$15.9          \\
            Human experts (post)\cite{camela2024dermatoscopic} & NA                       & 70.7$\pm$6.93 & 56.4$\pm$15.7          \\
            \cmidrule(lr){1-4}
            ResNet-18                                          & 0.735$\pm$0.023          & 70.0*         & 61.6$\pm$5.8          \\
            ResNet-50                                          & 0.755$\pm$0.018          & 70.0*         & 67.4$.\pm$4.3         \\
            BEiT-16-512                                        & \textbf{0.784$\pm$0.026} & 70.0*         & \textbf{70.0$\pm$6.6}  \\
            \bottomrule
        \end{tabular}
        \label{tab:results3}
    \end{table}

    The results in Table~\ref{tab:results3} offer several insights. At
    first glance, deep learning models appear to outperform human experts, with the BEiT
    model achieving a 14\% higher sensitivity compared to the human
    performance after the experts were instructed on the dermatoscopic predictors of BCC subtypes (70.0\% vs. 56.4\%).
    However, these comparisons should be considered preliminary, as the evaluation sets used in our experiments
    differ from those in the human expert study~\cite{camela2024dermatoscopic}. Within our
    dataset, the BEiT-based transformer notably surpassed the ResNet-based baselines,
    achieving the highest average AUC (0.784) and sensitivity (70.0\%).

    Despite the differences in evaluation conditions, these initial findings underscore
    the potential advantage of deep learning (and vision transformers in particular) for addressing the
    clinically challenging task
    of BCC subtyping from dermatoscopy images, thus highlighting their promise as future
    decision-support tools in dermatology practice.
    These findings are, however, subject to important limitations.
    The dataset originates from a single clinical center and is limited to Caucasian patients,
    which constrains the generalizability of our results to more diverse populations.
    Future work will aim to validate the proposed approach on multi-center, multi-ethnic
    cohorts and to establish more rigorous, controlled comparisons with human expert performance.

    \section{Conclusions}\label{sec:conclusions}
    We presented an initial proof-of-concept exploration of using deep learning to
    classify aggressive versus non-aggressive Basal Cell Carcinoma subtypes directly from dermatoscopic images.
    Our approach, based on vision transformers, achieved a respectable
    AUC of 0.784 in a dataset of 1271 BCC cases of various subtypes.
    Preliminary comparisons imply that deep
    learning models outperform human experts, although different dataset compositions preclude definitive
    conclusions. Nonetheless, our results suggest that deep learning holds much promise
    for non-invasive BCC subtyping, although additional work is needed to further reinforce these findings based on larger datasets and standardized evaluation protocols.

    \bibliographystyle{IEEEtran}
    \bibliography{main_abbrev}

\end{document}